\documentclass{article}

\usepackage{times}
\usepackage{graphicx} 
\usepackage{subfigure}

\usepackage{natbib}
\usepackage{macros}

\usepackage{algorithm}
\usepackage{algorithmic}

\usepackage{hyperref}
\usepackage{amssymb}
\usepackage{amsmath}
\usepackage{graphicx}
\usepackage{amsthm}
\usepackage{booktabs}
\usepackage[moderate,tracking=normal,wordspacing=normal,mathspacing=normal,mathdisplays=normal,margins=normal,indent=normal]{savetrees}
\theoremstyle{theorem}

\theoremstyle{definition}
\newtheorem{definition}{Definition}[section]

\theoremstyle{remark}
\newtheorem*{remark}{Remark}

\usepackage[accepted]{arxiv}

\usepackage{amsmath}

\icmltitlerunning{Noisy Activation Functions}

\begin{document}

\twocolumn[
\icmltitle{Noisy Activation Functions}

\icmlauthor{Caglar Gulcehre$^{\dagger\sharp}$}{gulcehrc@iro.umontreal.ca}
\icmlauthor{Marcin Moczulski$^{\dagger\diamond}$}{marcin.moczulski@stcatz.ox.ac.uk}
\icmlauthor{Misha Denil$^{\dagger\diamond}$}{misha.denil@gmail.com}
\icmlauthor{Yoshua Bengio$^{\dagger\sharp}$}{bengioy@iro.umontreal.ca}
\icmladdress{$^{\dagger\sharp}$University of Montreal}
\icmladdress{$^{\dagger\diamond}$University of Oxford}

\icmlkeywords{deep learning, machine learning, noisy activations, activation functions, machine
translation, image caption generation, LSTMs}

]

\begin{abstract}
  Common nonlinear activation functions used in neural networks can cause training difficulties due to 
  the saturation behavior of the activation function, which may hide dependencies 
  that are not visible to vanilla-SGD (using first order gradients only). Gating mechanisms 
  that use softly saturating activation functions to emulate the discrete switching 
  of digital logic circuits are good examples of this. We propose to exploit the injection of
  appropriate noise so that the gradients may flow easily, even if the noiseless 
  application of the activation function would yield zero gradient.  Large noise 
  will dominate the noise-free gradient and allow stochastic gradient descent to
  explore more. By adding noise only to the problematic parts of the activation 
  function, we allow the optimization procedure to explore the boundary between the degenerate (saturating) and
  the well-behaved parts of the activation function. We also establish connections to simulated annealing,
  when the amount of noise is annealed down, making it easier to optimize hard objective functions.
  We find experimentally that replacing such saturating activation functions by noisy variants helps training in
  many contexts, yielding state-of-the-art or competitive results on different datasets and task, especially
  when training seems to be the most difficult, e.g., when curriculum learning is necessary to obtain good results.
\end{abstract}

\section{Introduction}

The introduction of the piecewise-linear activation functions such as ReLU and Maxout 
\cite{goodfellow2013maxout}  units had a profound effect on deep learning, and was a 
major catalyst in allowing the training of much deeper networks.
It is thanks to ReLU that for the first time it was shown~\citep{glorot2011deep}
that deep purely supervised networks can be trained, whereas using 
$\tanh$ nonlinearity only allowed to train shallow networks.
A plausible hypothesis about the recent surge of interest on these piecewise-linear activation functions \citep{glorot2011deep},
is due to the fact that they are easier to optimize with SGD and backpropagation than smooth
activation functions, such as $\sigm$ and $\tanh$. The recent successes of piecewise linear functions 
is particularly evident in computer vision, where the ReLU
has become the default choice in convolutional networks.

We propose a new technique to train neural networks with activation functions which 
strongly saturate when their input is large. This is mainly achieved by injecting noise to the 
activation function in its saturated regime and learning the level of noise. Using this 
approach, we have found that it was possible to train neural networks with much wider 
family of activation functions than previously. Adding noise to the activation function has been considered for ReLU units and was 
explored in \cite{bengio2013estimating, nair2010rectified} for feed-forward networks 
and Boltzmann machines to encourage units to explore more and make the optimization easier.

More recently there has been a resurgence of interest in more elaborated ``gated'' 
architectures such as LSTMs \cite{hochreiter1997long} and GRUs \cite{cho2014learning}, 
but also encompassing neural attention mechanisms that have been used in the NTM 
\cite{graves2014neural}, Memory Networks \cite{weston2014memory}, automatic image 
captioning \cite{xu2015show}, video caption generation \cite{yao2015capgenvid} 
and wide areas of applications \cite{lecun2015deep}. A common thread running 
through these works is the use of soft-saturating non-linearities, such as the 
sigmoid or softmax, to emulate the hard decisions of digital logic circuits. In 
spite of its success, there are two key problems with this approach.

\begin{enumerate}
\item Since the non-linearities still saturate there are problems with vanishing gradient information flowing through the gates; and
\item since the non-linearities only \emph{softly} saturate they do not allow one to take hard decisions.
\end{enumerate}
Although gates often operate in the soft-saturated regime
\cite{karpathy2015visualizing,bahdanau2014neural,hermann2015teaching} the architecture prevents them
from being fully open or closed.  We follow a novel approach to address both of these problems.
Our method addresses the second problem through the use of hard-saturating nonlinearities, which allow gates to make perfectly
on or off decisions when they saturate.  Since the gates are able to be completely open or closed, no information is lost
through the leakiness of the soft-gating architecture.

By introducing hard-saturating nonlinearities, we have exacerbated the problem of gradient flow, since gradients in the
saturated regime are now precisely zero instead of being negligible. However, by introducing 
noise into the activation function which can grow based on the magnitude of saturation, we 
encourage random exploration. 

At test time the noise in the activation functions can be removed or replaced with the expectation, 
and as our experiments show, the resulting deterministic networks outperform their 
soft-saturating counterparts on a wide variety of tasks, and allow to reach state-of-the-art performance
by simple drop-in replacement of the nonlinearities in existing training code.

The technique that we propose, addresses the difficulty of optimization 
and having hard-activations at the test time for gating units and we propose a way of performing 
simulated annealing for neural networks.

\citet{hannun2014deep,le2015simple} used ReLU activation functions with simple RNNs. In this paper,
we successfully show that, it is possible to use piecewise-linear activation functions 
with gated recurrent networks such as LSTM and GRU's.

\section{Saturating Activation Functions}
\label{sec:activation-fn}

\begin{definition}{\textbf{(Activation Function)}}.
    An activation function is a function $\h : \R \to \R$ that is differentiable almost everywhere.
\end{definition}

\vspace*{-2mm}
\begin{definition}{\textbf{(Saturation)}}.
    An activation function $\h(x)$ with derivative $\h^{\prime}(x)$ is 
    said to right (resp.\ left) saturate if its limit as $x\to\infty$ (resp.\ $x\to-\infty$)
    is zero.  An activation function is said to saturate (without qualification) 
    if it both left and right saturates.
\end{definition}
\vspace*{-2mm}

Most common activation functions used in recurrent networks (for example, $\tanh$ 
and $\sigm$) are saturating.  In particular they are soft saturating, meaning that 
they achieve saturation only in the limit.

\begin{definition}{\textbf{(Hard and Soft Saturation)}}.
    Let $c$ be a constant such that $x > c$ implies $\h^{\prime}(x) = 0$ and left 
hard saturates when $x < c$ implies $\h^{\prime}(x) = 0$, $\forall x$. We say that 
$\h(\cdot)$ hard saturates (without qualification) if it both left and right hard saturates. 
An activation function that saturates but achieves zero gradient only in the limit 
is said to soft saturate.
\end{definition}

We can construct hard saturating versions of soft saturating activation functions 
by taking a first-order Taylor expansion about zero and clipping the results to an 
appropriate range.

For example, expanding $\tanh$ and $\sigm$ around $0$, with $x\approx 0$,
we obtain linearized functions $\fu^t$ and $\fu^s$ of $\tanh$ and $\sigm$ respectively:
\begin{align}
  \label{eq:linearizations}
    \sigm(x) & \approx  \fu^s(x) = 0.25 x + 0.5 \\
    \tanh(x) & \approx \fu^t(x) = x.
\end{align}
Clipping the linear approximations result to,
\begin{align}
	\hardsigm(x) = \max(\min(\fu^s(x),~1),~0) \\
    \hardtanh(x) = \max(\min(\fu^t(x),~1),~-1).
\end{align}

The motivation behind this construction is to introduce linear behavior around zero to allow 
gradients to flow easily when the unit is not saturated, while providing a crisp decision 
in the saturated regime.

\begin{figure}[ht]
\includegraphics[width=0.95\columnwidth]{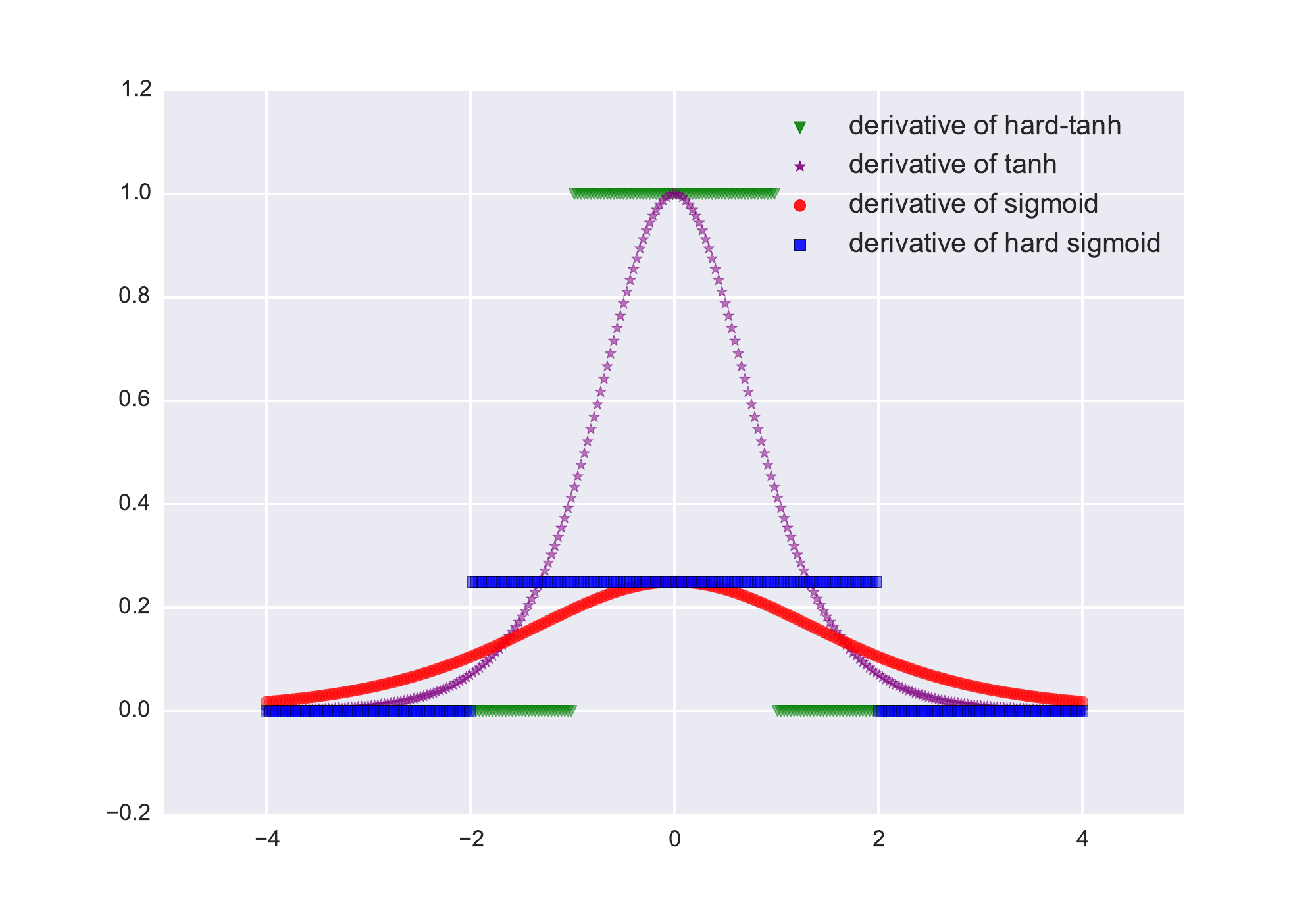}
\caption{The plot for derivatives of different activation functions.}
\centering
\end{figure}

The ability of the hard-sigmoid and hard-tanh to make crisp decisions comes 
at the cost of exactly 0 gradients in the saturated regime. This can cause 
difficulties during training: a small but not infinitesimal change of the 
pre-activation (before the nonlinearity) may help to reduce the 
objective function, but this will not be reflected in the gradient.

In the rest of the document we will use $h(x)$ to refer to a 
generic activation function and use $u(x)$ to denote its
linearization based on the first-order Taylor expansion about zero.
$\hardsigm$ saturates when $x \leqslant -2$ or $x \geqslant 2$
and $\hardtanh$  saturates when $x \leqslant -1$ or $x \geqslant 1$.
We denote the threshold by $x_t$. Absolute values of the threshold are $x_t=2$ for $\hardsigm$
and $x_t=1$ for the $\hardtanh$.

\begin{remark}
    Let us note that both $\hardsigm(x)$, $\sigm(x)$  and $\tanh(x)$ are all contractive mapping.
    $\hardtanh(x)$, becomes a contractive mapping only when its input is greater than the threshold.
    An important difference among these activation functions is their fixed points. 
    $\hardsigm(x)$ has a fixed point at $x=\frac{2}{3}$. However the fixed-point of $\sigm$ is $x\approx0.69$. 
    Any $x \in \R$ between $-1$ and $1$ can be the fixed-point of $\hardtanh(x)$, 
    but the fixed-point of $\tanh(x)$ is $0$. $\tanh(x)$ and $\sigm(x)$ have point attractors at their fixed-points. 
    Those mathematical differences among the saturating activation functions 
    can make them behave differently with RNNs and deep networks.
\end{remark}

The highly non-smooth gradient descent trajectory may bring
the parameters into a state that pushes the activations of 
a unit towards the $0$ gradient regime for a particular example, 
from where it may become difficult to escape and the unit 
may get stuck in the $0$ gradient regime.

When units saturate and gradients vanish, an algorithm may require many training
examples and a lot of computation to recover.

\section{Annealing with Noisy Activation Functions}

Consider a noisy activation function $\phi(x,\xi)$ in which we have injected iid noise $\xi$,
to replace a saturating nonlinearity such as the $\hardsigm$ and $\hardtanh$
introduced in the previous section. In the next section we describe the proposed
noisy activation function which has been used for our experiments, but here we
want to consider a larger family of such noisy activation functions, when we
use a variant of stochastic gradient descent (SGD) for training.

Let $\xi$ have variance $\sigma^2$ and mean $0$. We want to characterize what happens
as we gradually anneal the noise, going from large noise levels ($\sigma\rightarrow \infty$) to
no noise at all $(\sigma \rightarrow 0)$.

Furthermore, we will assume that $\phi$ is such that when the noise level
becomes large, so does its derivative with respect to $x$:
\begin{equation}
  \label{eq:large-xi-derivative}
  \lim_{|\xi|\rightarrow\infty} |\frac{\partial \phi(x,\xi)}{\partial x}| \rightarrow \infty.
  \end{equation}
\begin{figure}[htp]
\centerline{\resizebox{0.4\textwidth}{!}{\includegraphics{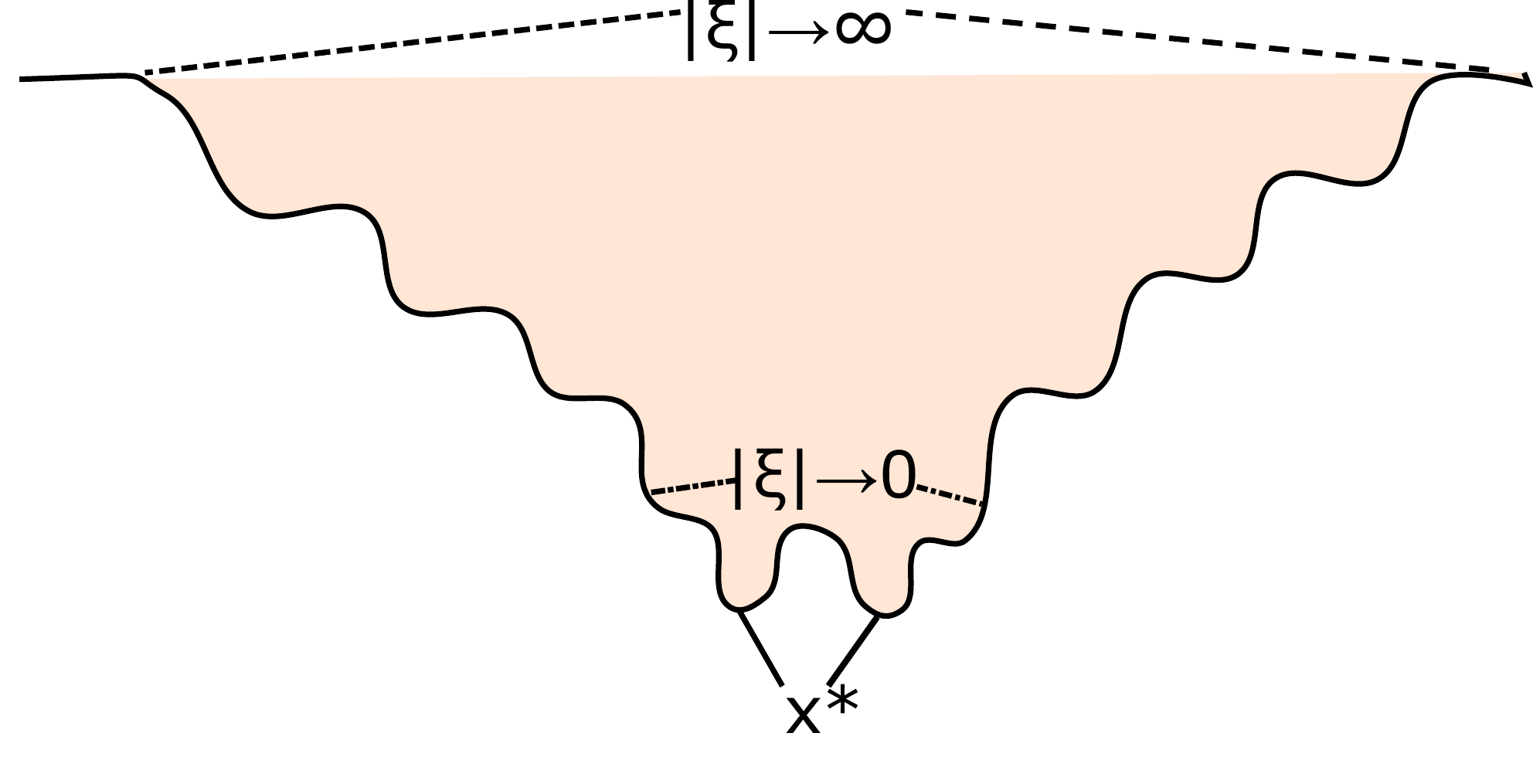}}}
\caption{\sl An example of a one-dimensional, non-convex objective function where a simple 
    gradient descent will behave poorly. With large noise $|\xi| \rightarrow \infty$, SGD can escape from 
    saddle points and bad local-minima as a result of exploration. As we anneal the noise level $|\xi| \rightarrow 0$, 
    SGD will eventually converge to one of the local-minima $x^{\ast}$\label{fig:nonconvex_noise}.
}
\end{figure}

In the $0$ noise limit, we recover a deterministic nonlinearity, $\phi(x,0)$,
which in our experiments is piecewise linear and allows us to capture the kind of
complex function we want to learn. As illustrated in Figure~\ref{fig:nonconvex_noise}, in the large noise limit, 
large gradients are obtained because backpropagating through $\phi$ gives rise to
large derivatives. Hence, the noise drowns the signal:
the example-wise gradient on parameters is much larger than it would have been with $\sigma=0$.
SGD therefore just sees noise and can move around anywhere in parameter space without ``seeing''
any trend.

Annealing is also related to the signal to noise ratio where $SNR$ can be defined as the ratio of 
the variance of noise $\sigma_{\text{signal}}$ and $\sigma_{\text{noise}}$, 
$SNR=\frac{\sigma_{\text{signal}}}{\sigma_{\text{noise}}}$. If $SNR \rightarrow 0$, the model will 
do pure random exploration. As we anneal $SNR$ will increase, and
when $\sigma_{\text{noise}}$ converges to $0$, the only source of
exploration during the training will come from the noise of Monte Carlo 
estimates of stochastic gradients.

This is precisely what we need for methods such as simulated annealing~\citep{Kirkpatrick83} and
continuation methods~\citep{Allgower80-short} to be helpful, in the context of the optimization of
difficult non-convex objectives. With high noise, SGD is free to explore all parts of space.
As the noise level is decreased, it will prefer some regions where the signal is strong enough
to be ``visible'' by SGD: given a finite number of SGD steps, the noise is not averaged out,
and the variance continues to dominate. Then as the noise level is reduced SGD spends more time
in ``globally better'' regions of parameter space. As it approaches to zero we are fine-tuning
the solution and converging near a minimum of the noise-free objective function. A related approach
of adding noise to gradients and annealing the noise was investigated in \cite{neelakantan2015adding}
as well. \citet{ge2015escaping} showed that SGD with annealed noise will globally converge 
to a local-minima for non-convex objective functions in polynomial number of iterations. Recently, 
\citet{mobahi2016training} propose an optimization method that applies Gaussian smoothing 
on the loss function such that annealing weight noise is a Monte Carlo estimator of that.

\vspace*{-1mm}
\section{Adding Noise when the Unit Saturate}

A novel idea behind the proposed noisy activation is that {\bf the amount of noise 
added to the nonlinearity is proportional to the magnitude of saturation of the nonlinearity}. 
For $\hardsigm(x)$ and $\hardtanh(x)$, due to our parametrization of the noise, 
that translates into the fact that the noise is only added when the hard-nonlinearity 
saturates. This is different from previous proposals such as the noisy rectifier
from~\citet{Bengio-arxiv2013} where noise is added
just before a rectifier (ReLU) unit, independently of whether the input
is in the linear regime or in the saturating regime of the nonlinearity. 

The motivation is to keep the training signal clean when the unit is in the
non-saturating (typically linear) regime and provide some noisy signal when
the unit is in the saturating regime.

$\h(x)$ refer to hard saturation activation function such as the hard-sigmoid and hard-tanh
introduced in Sec.~\ref{sec:activation-fn},
we consider noisy activation functions of the following form:
\begin{equation}
    \phi(x,\xi) = \h(x) + s  
\end{equation}
and $s = \mu + \sigma \xi$.  Here $\xi$ is an iid random variable
drawn from some generating distribution, and the parameters $\mu$ and
$\sigma$ (discussed below) are used to generate a location scale family
from $\xi$.

Intuitively when the unit saturates we pin its output to the threshold value $t$ and add noise.
The exact behavior of the method depends on the type of noise $\xi$ and the choice
of $\mu$ and $\sigma$, which we can pick as functions of $x$ in order to let some
gradients be propagated even when we are in the saturating regime.

A desirable property we would like $\phi$ to approximately satisfy is that,
in expectation, it is equal to the hard-saturating activation function, i.e.
\begin{align}
    \E_{\xi \sim \mathcal{N}(0,1)}[\phi(x,\xi)] \approx \h(x)
  \label{eq:expectation-condition}
\end{align}
If the $\xi$ distribution has zero mean then this property can be
satisfied by setting $\mu=0$, but for biased noise it will be necessary to
make other choices for $\mu$. In practice, we used slightly biased $\phi$
with good results.

Intuitively we would like to add more noise when $x$ is far into the
saturated regime, since a large change in parameters would be required
desaturate $h$.  Conversely, when $x$ is close to the saturation threshold
a small change in parameters would be sufficient for it to escape.  To that
end we make use of the difference between the original activation function $h$
and its linearization $u$
\begin{align}
  \Delta = \h(x) - \fu(x)
\end{align}

when choosing the scale of the noise. See~Eqs.\ref{eq:linearizations} for definitions
of $u$ for the $\hardsigm$ and $\hardtanh$ respectively. The quantity $\Delta$ is zero in the unsaturated regime,
and when $h$ saturates it grows proportionally to the distance between $|x|$ and the saturation threshold $x_t$.
We also refer $|\Delta|$ as the magnitude of the saturation. 

We experimented with different ways of scaling $\sigma$ with $\Delta$,
and empirically found that the following formulation performs better:
\begin{equation}
    \label{eqn:parameterization_std}
    \begin{aligned}
        & \sigma(x) = c~(\g(p \Delta) - 0.5)^2 \\ 
        & \g(x) = \text{sigmoid}(x) .
    \end{aligned}
\end{equation} \\
In Equation~\ref{eqn:parameterization_std} a free scalar parameter $p$ is learned during the course of training.
By changing $p$, the model is able to adjust the magnitude of the noise and that also effects the
sign of the gradient as well. The hyper-parameter $c$ changes the scale of the standard deviation of the noise.

\subsection{Derivatives in the Saturated Regime}
\vspace*{-1mm}

In the simplest case of our method we draw $\xi$ from an unbiased distribution, such as a standard normal.
In this case we choose $\mu=0$ to satisfy Equation~\ref{eq:expectation-condition} and therefore we
will have,
\begin{align*}
  \phi(x,\xi) =  \h(x) + \sigma(x) \xi  
\end{align*}

\begin{figure}[t]
    \label{fig:noisy_tanh_gauss}
	\begin{center}
		\includegraphics[scale=0.4]{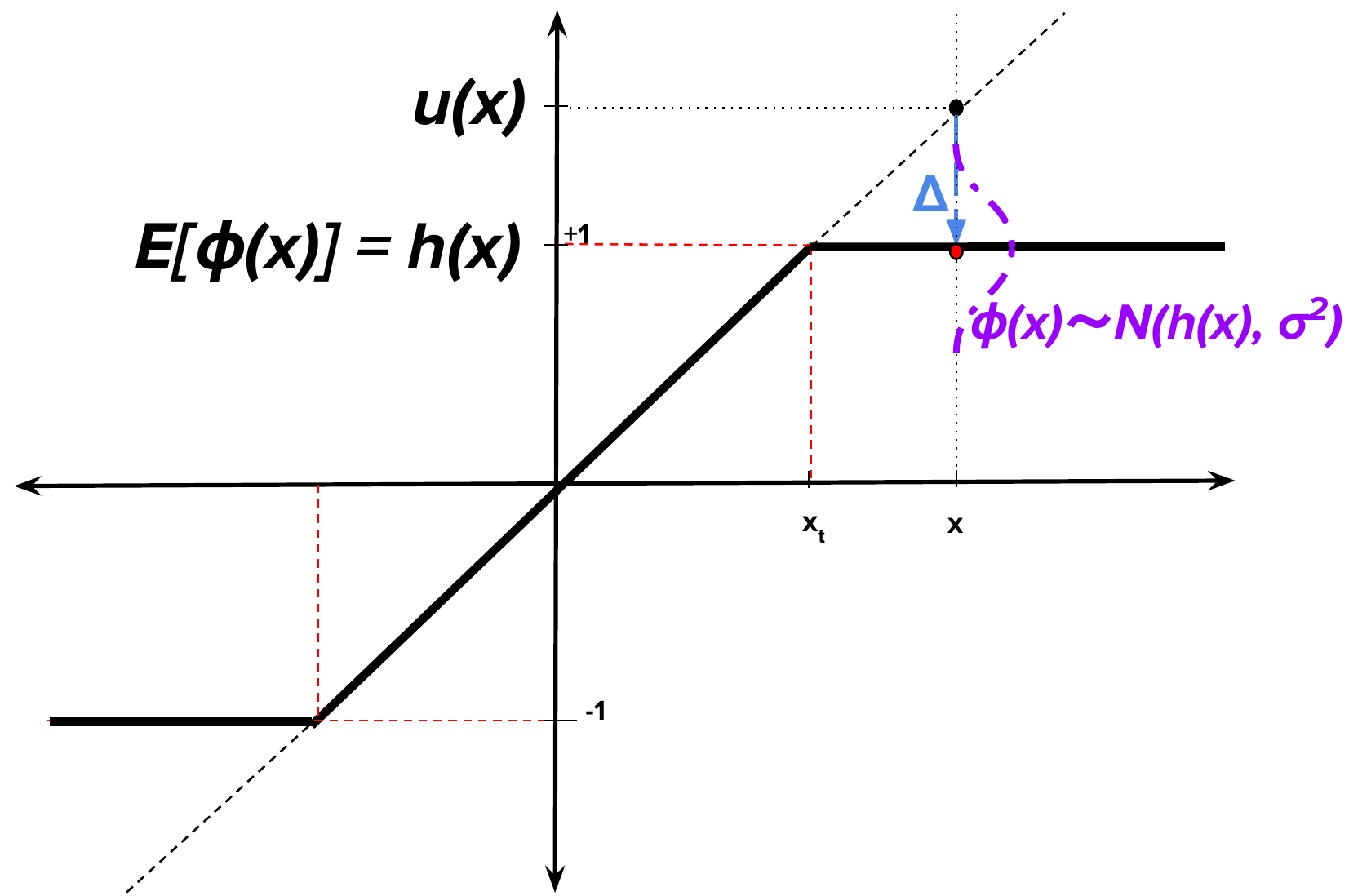}
		\caption{A simple depiction of adding Gaussian noise on the linearized activation function, which
      brings the average back to the hard-saturating nonlinearity $\h(x)$, in bold. Its linearization
      is $\fu(x)$ and the noisy activation is $\phi$. The difference $\h(x)-\fu(x)$ is $\Delta$ which is a vector
      indicates the discrepancy between the linearized function and the actual function that
      the noise is being added to $\h(x)$. Note that, $\Delta$ will be zero, at the
      non-saturating parts of the function where $\fu(x)$ and $\h(u)$ matches perfectly.}
	\end{center}
\end{figure}

Due to our parameterization of $\sigma(x)$, when $|x| \le x_t$ our stochastic 
activation function behaves exactly as the linear function $\fu(x)$, 
leading to familiar territory. Because $\Delta$ will be $0$. Let us for the 
moment restrict our attention to the case when $|x| > x_t$ and $h$ saturates. 
In this case the derivative of $\h(x)$ is precisely zero, however, 
if we condition on the sample $\xi$ we have
\begin{align}
  \phi'(x, \xi) = \frac{\partial}{\partial x} \phi(x, \xi) = \sigma'(x)\xi
  \label{eq:gradient-flow}
\end{align}
which is non-zero almost surely.

In the non-saturated regime, where $\phi^{\prime}(x, \xi) = \h^{\prime}(x)$ the
optimization can exploit the linear structure of $h$ near the origin in
order to tune its output.  In the saturated regime the randomness in $\xi$
drives exploration, and gradients still flow back to $x$ since the scale of
the noise still depends on $x$.  To reiterate, we get gradient information
at every point in spite of the saturation of $\h$, and the variance of the 
gradient information in the saturated regime depends on the variance 
of $\sigma^{\prime}(x)\xi$.

\subsection{Pushing Activations towards Linear Regime}

An unsatisfying aspect of the formulation with unbiased noise is that,
depending on the value of $\xi$ occasionally the gradient of $\phi$ will point
the wrong way.  This can cause a backwards message that would push $x$ in
a direction that would worsen the objective function on average over $\xi$.
Intuitively we would prefer these messages to
``push back'' the saturated unit towards a non-saturated state where the
gradient of $\h(x)$ can be used safely.

A simple way to achieve this is to make sure that the noise $\xi$ is always
positive and adjust its sign manually to match the sign of $x$.  In
particular we could set
\begin{align*}
    \fd(x) =& -\operatorname{sgn}(x) \operatorname{sgn}(1 - \alpha) \\
    s =& \mu(x) + \fd(x)\sigma(x) |\xi|.
\end{align*}
where $\xi$ and $\sigma$ are as before and $\sgn$ is the sign function, 
such that $\sgn(x)$ is $1$ if $x$ is greater than or equal to $0$ otherwise it is $-1$.
We also use the absolute value of $\xi$ in the reparametrization of the noise, 
such that the noise is being sampled from a half-Normal distribution. 
We ignored the sign of $\xi$, such that the direction that the noise pushes the activations
are determined by $\fd(x)$ and it will point towards $\h(x)$. Matching the sign of 
the noise to the sign of $x$ would ensure that we avoid the sign 
cancellation between the noise and the gradient message from backpropagation.
$\sgn(1 - \alpha)$ is required to  push the activations towards $\h(x)$ when 
the bias from $\alpha$ is introduced.

\begin{figure}[ht]
    \centering
    \includegraphics[width=0.99\columnwidth]{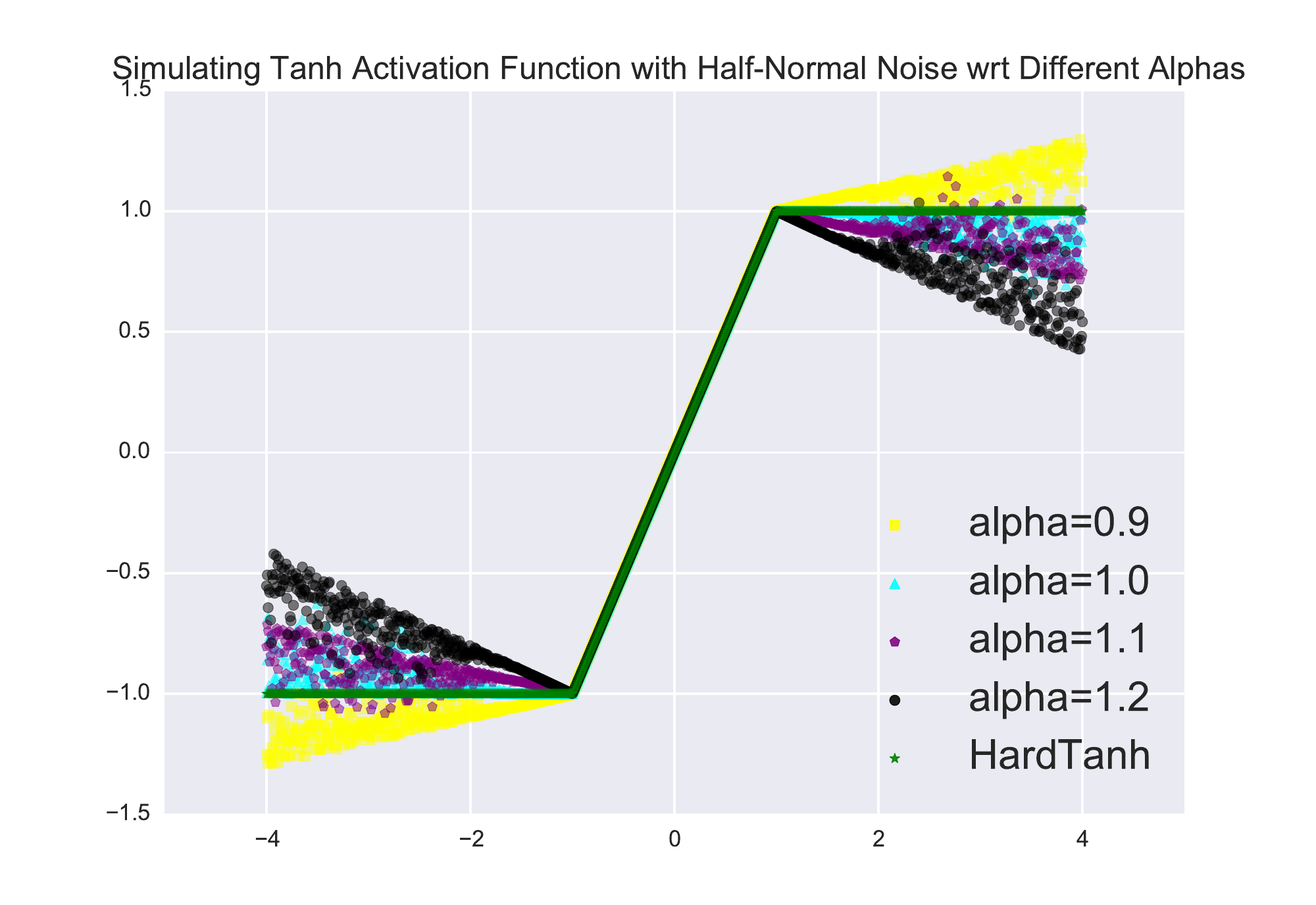}
    \caption{Stochastic behavior of the proposed noisy  activation function with
        different $\alpha$ values and with noise sampled from the Normal 
        distribution, approximating the $\hardtanh$ nonlinearity (in bold green).}
    \label{fig:hard_tanh_sim_gauss}
\end{figure}

In practice we use a hyperparameter $\alpha$ that influences the mean of the added term,
such that $\alpha$ near 1 approximately satisfies the above condition, as seen
in Fig.~\ref{fig:hard_tanh_sim_gauss}.
We can rewrite the noisy term $s$ in a way that the noise can either be added to the linearized
function or $\h(x)$. The relationship between $\Delta$, $\fu(x)$ and $\h(x)$ is visualized 
Figure~\ref{fig:noisy_tanh_gauss} can be expressed as in Eqn~\ref{eqn:theoretical_ver}.

We have experimented with different types of noise. Empirically, in terms of performance
we found, half-normal and normal noise to be better. In Eqn~\ref{eqn:theoretical_ver}, we provide the formulation for 
the activation function where $\epsilon=|\xi|$ if the noise is sampled from half-normal distribution,
$\epsilon=\xi$ if the noise is sampled from normal distribution.

\begin{equation}
\label{eqn:theoretical_ver}
\begin{aligned}
    \phi(x, \xi) &= \fu(x) + \alpha \Delta + \fd(x) \sigma(x) \epsilon \\
\end{aligned}
\end{equation}

By using Eqn~\ref{eqn:theoretical_ver}, we arrive at the noisy activations,
which we used in  our experiments.
\begin{equation}
    \label{eqn:experimental_ver}
    \phi(x, \xi) = \alpha \h(x) + (1 - \alpha) \fu(x) + \fd(x) \sigma(x) \epsilon
\end{equation}

As can be seen in Eqn~\ref{eqn:experimental_ver}, there are three paths that gradients
can flow through the neural network, the linear path ($\fu(x)$), nonlinear path ($\h(x)$) and the stochastic 
path ($\sigma(x)$). The flow of gradients through these different pathways across different layers 
makes the optimization of our activation function easier.

At test time, we used the expectation of Eqn~\ref{eqn:experimental_ver} in order to get deterministic
units,
\begin{equation}
    \label{eqn:test_ver}
    \E_{\xi}[\phi(x, \xi)] = \alpha \h(x) + (1 - \alpha) \fu(x) + \fd(x) \sigma(x) \E_{\xi}[\epsilon]
\end{equation}

If $\epsilon=\xi$, then $\E_{\xi}[\epsilon]$ is $0$. Otherwise if
$\epsilon=|\xi|$, then $\E_{\xi}[\epsilon]$ is $\sqrt{\frac{2}{\pi}}$.

\begin{algorithm}
    \caption{Noisy Activations with Half-Normal Noise for Hard-Saturating Functions}
    \label{algo:moll_fn}
    \begin{algorithmic}[1]
\STATE $\Delta \gets h(x) - u(x)$
\STATE $\fd(x) \gets - \sgn(x) \operatorname{sgn}(1 - \alpha)$
\STATE $\sigma(x) \gets c~(g(p \Delta) - 0.5)^2$
\STATE $\xi \sim \mathcal{N}(0,~1)$
\STATE $\phi(x, \xi) \gets \alpha \h(x) + (1 - \alpha) \fu(x) + ( \fd(x) \sigma(x) |\xi|)$
\end{algorithmic}
\end{algorithm}

To illustrate the effect of $\alpha$ and noisy activation of the $\hardtanh$,
We provide plots of our stochastic activation functions in Fig~\ref{fig:hard_tanh_sim_gauss}. 

\section{Adding Noise to Input of the Function}

Adding noise with fixed standard deviation to the input of the activation function 
has been investigated for ReLU activation functions \cite{nair2010rectified,bengio2013estimating}.

\begin{equation}
    \label{eqn:noise_at_inp}
    \phi(x, \xi) = \h(x + \sigma \xi)~\text{and}~\xi \sim \mathcal{N}(0, 1).
\end{equation}

In Eqn~\ref{eqn:noise_at_inp}, we provide a parametrization of the noisy activation function.
$\sigma$ can be either learned as in Eqn~\ref{eqn:parameterization_std} or fixed as a hyperparameter.

The condition in Eqn~\ref{eq:large-xi-derivative} is satisfied only when $\sigma$ is learned. Experimentally we found
small values of $\sigma$ to work better. When $\sigma$ is fixed and small, as $x$ gets larger 
and further away from the threshold $x_t$, noise will less likely be able to
push the activations back to the linear regime. We also investigated the effect of injecting input noise
when the activations saturate:
\begin{equation}
    \label{eqn:noise_at_inp_sat}
    \phi(x, \xi) = \h(x + \mathbf{1}_{|x| \ge |x_t|}(\sigma \xi))~\text{and}~\xi \sim \mathcal{N}(0, 1).
\end{equation}

\section{Experimental Results}
In our experiments, we used noise only during training: at test time we replaced 
the noise variable with its expected value. We performed our experiments with 
just a drop-in replacement of the activation functions in existing experimental 
setups, without changing the previously set hyper-parameters. Hence it is plausible 
one could obtain better results by performing a careful hyper-parameter tuning 
for the models with noisy activation functions. In all our experiments, 
we initialized $p$ uniform randomly from the range $\left[-1,~1\right]$.

We provide experimental results using noisy activations with normal \textbf{(NAN)}, half-normal noise 
\textbf{(NAH)}, normal noise at the input of the function \textbf{(NANI)},
normal noise at the input of the function with learned $\sigma$ \textbf{(NANIL)} and 
normal noise injected to the input of the function when the unit saturates \textbf{(NANIS)}.
Codes for different types of noisy activation functions can be found at \url{https://xx}.

\subsection{Exploratory Analysis}
\begin{figure}[h]
    \centering
    \includegraphics[width=0.95\columnwidth]{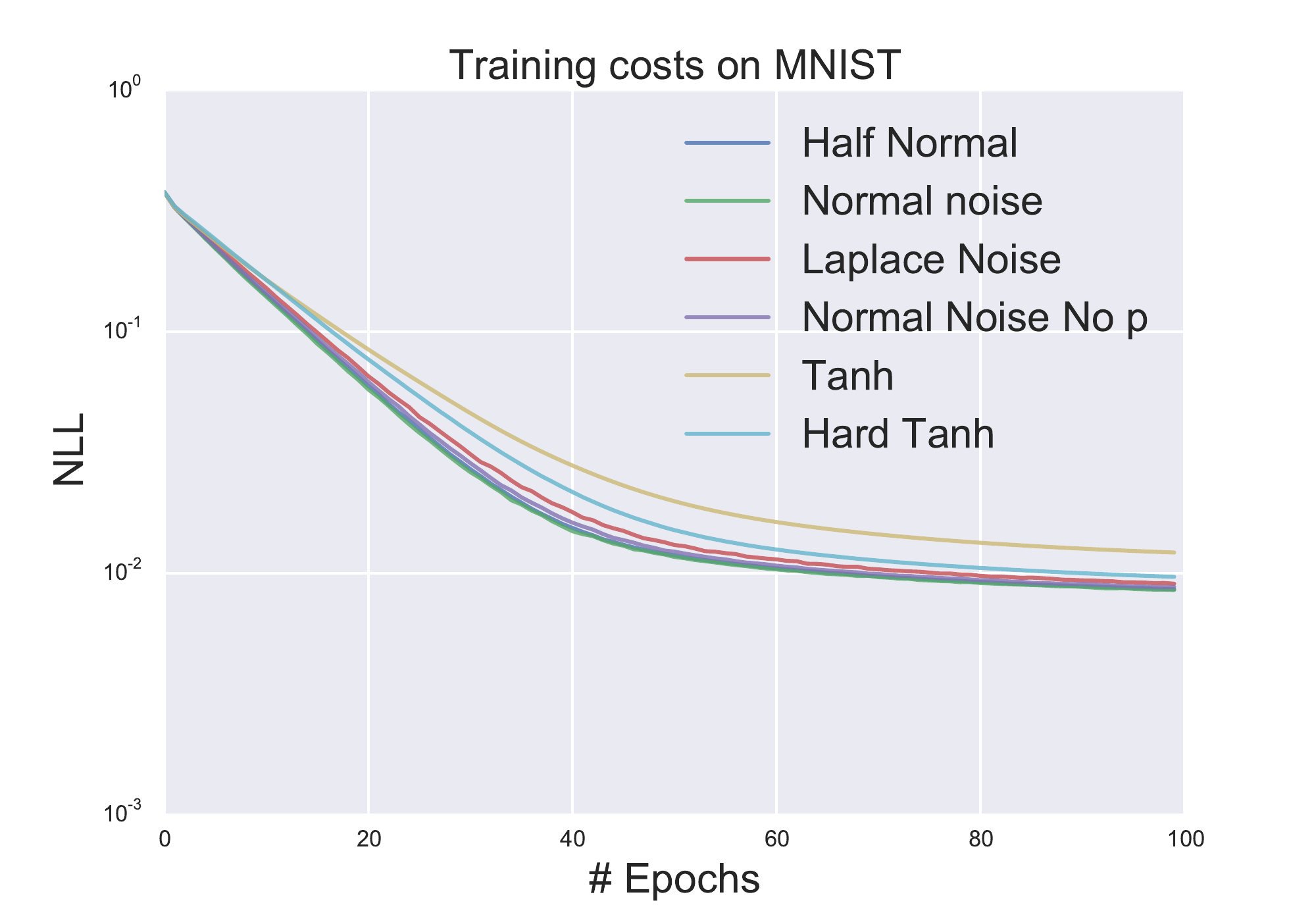}
    \caption{Learning curves of a single layer MLP trained with RMSProp with different noise types
    and activation functions}
    \label{fig:mnist_different_noise}
\end{figure}

As a sanity-check, we performed small-scale control experiments, in order to 
observe the behavior of the noisy units. In  Fig~\ref{fig:mnist_different_noise}, 
we showed the learning curves of different types of activations with various types 
of noise in contrast to the $\tanh$ and $\hardtanh$ units. The models are single-layer 
MLPs trained on MNIST for classification and we show the average negative
log-likelihood $-\log P({\rm correct\; class} |{\rm input})$. In general, we found that 
models with noisy activations converge faster than those using $\tanh$ and 
$\hardtanh$ activation functions, and to lower NLL than the $\tanh$ network.

\begin{figure}[h]
    \centering
    \includegraphics[width=0.95\columnwidth]{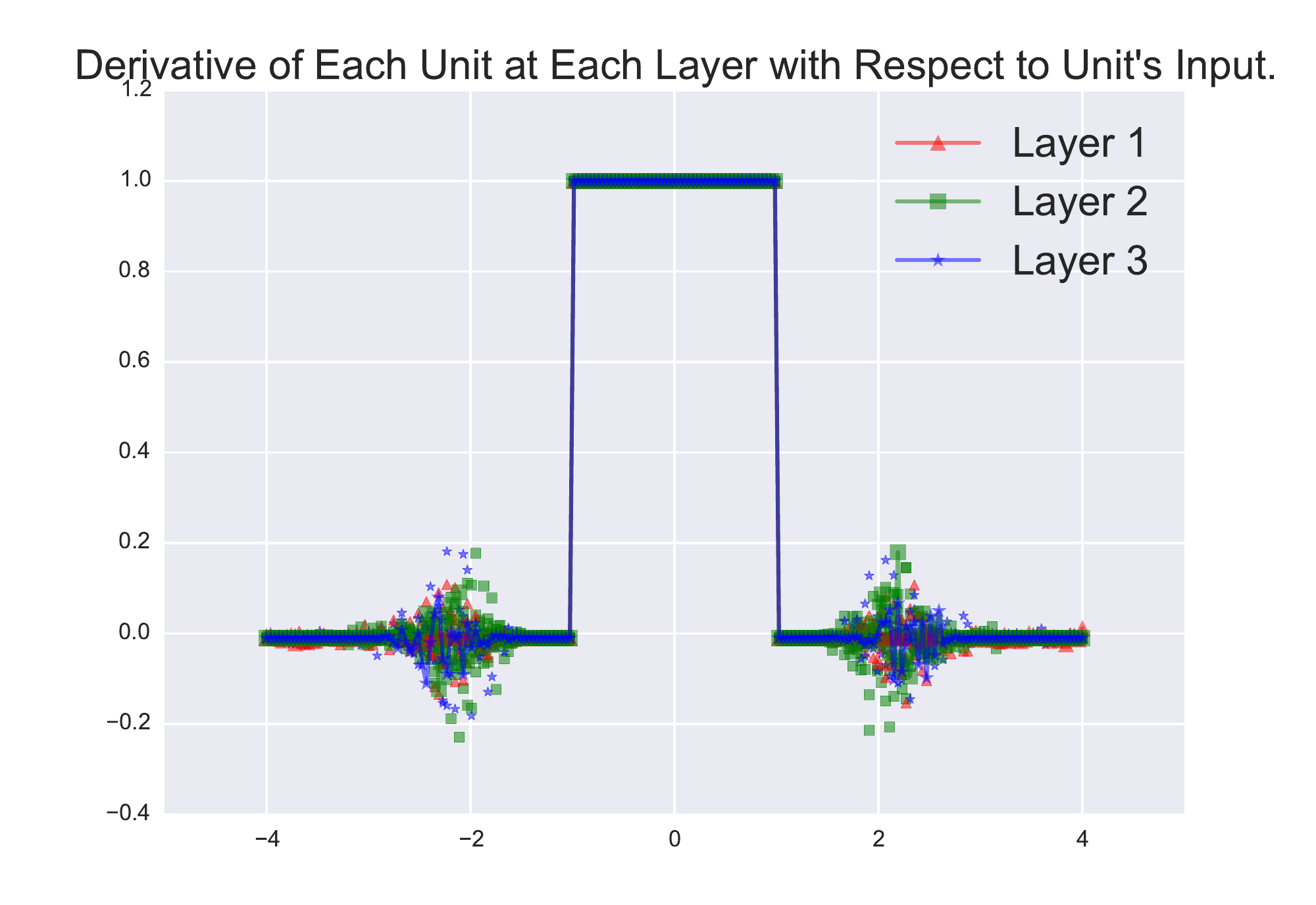}
    \caption{Derivatives of each unit at each layer with respect to its input for a three-layered
    MLP trained on a dataset generated by three normal distributions with different means and standard
    deviations. In other words learned $\frac{\partial \phi(x^k_i, \xi^k_i)}{\partial x^k_i}$ at the end
    of training for $i^{th}$ unit at $k^{th}$ layer. $\xi^k$ is sampled from Normal distribution
    with $\alpha=1$.}
    \label{fig:derivlayerwrtInp}
\end{figure}

\begin{figure}[h]
    \centering
    \includegraphics[width=0.95\columnwidth]{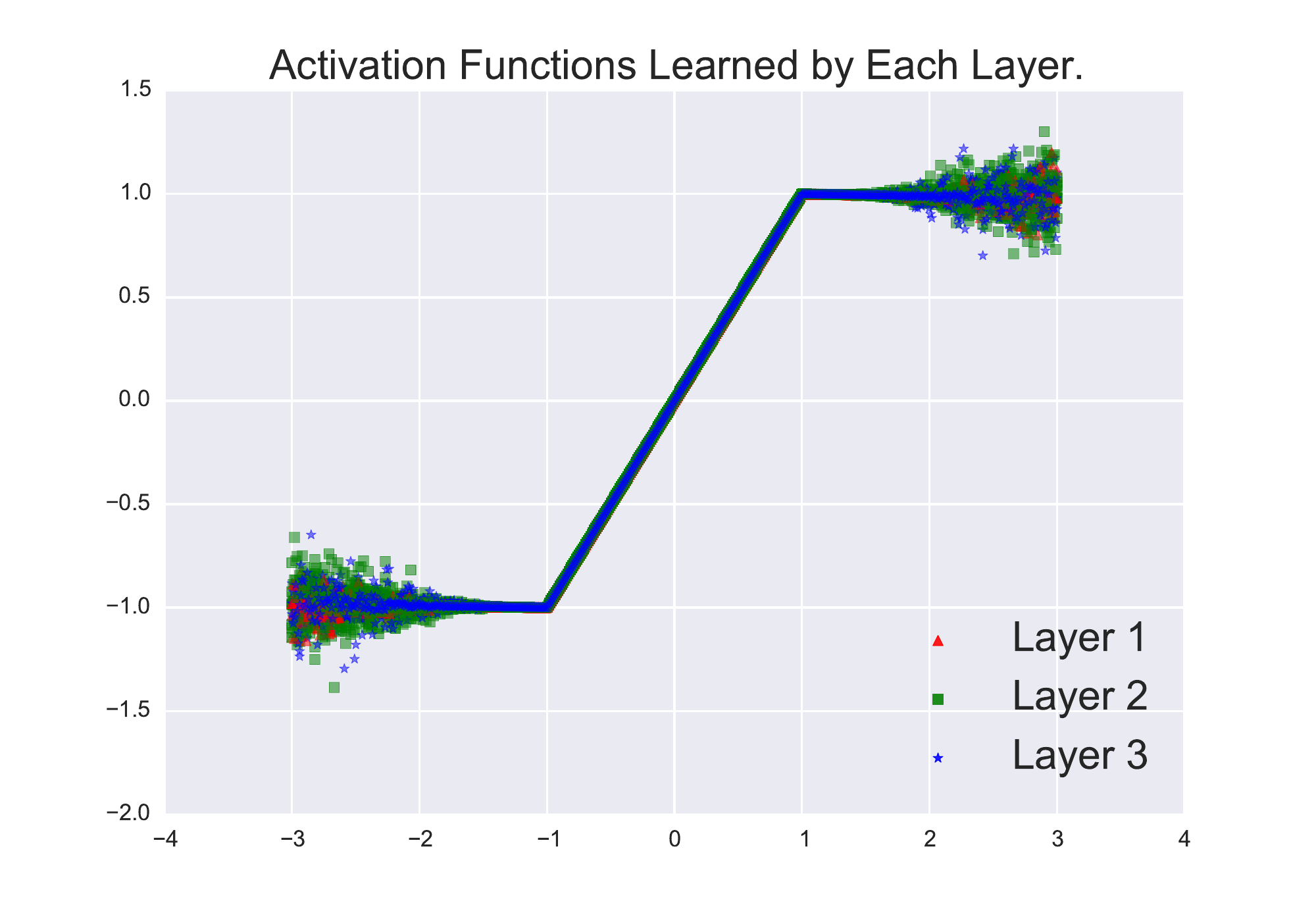}
    \caption{Activations of each unit at each layer of a three-layer
    MLP trained on a dataset generated by three normal distributions with different means and standard
    deviations. In other words learned $\phi(x^k_i, \xi^k_i)$ at the end of training for
    $i^{th}$ unit at $k^{th}$ layer. $\xi^k$ is sampled from Half-Normal distribution  with
    $\alpha=1$.}
    \label{fig:actLayerwise}
\end{figure}

We trained $3-$layer MLP on a dataset generated from a mixture of $3$ Gaussian distributions
with different means and standard deviations. Each layer of the MLP contains $8$-hidden units.
Both the model with $\tanh$ and noisy$-\tanh$ activations was able to solve this task almost
perfectly. By using the learned $p$ values, in Figure~\ref{fig:derivlayerwrtInp}
and~\ref{fig:actLayerwise}, we showed the scatter plot of the activations of each unit
at each layer and the derivative function of each unit at each layer with respect to
its input.

\begin{figure}[htp]
    \centering
    \includegraphics[width=0.95\columnwidth]{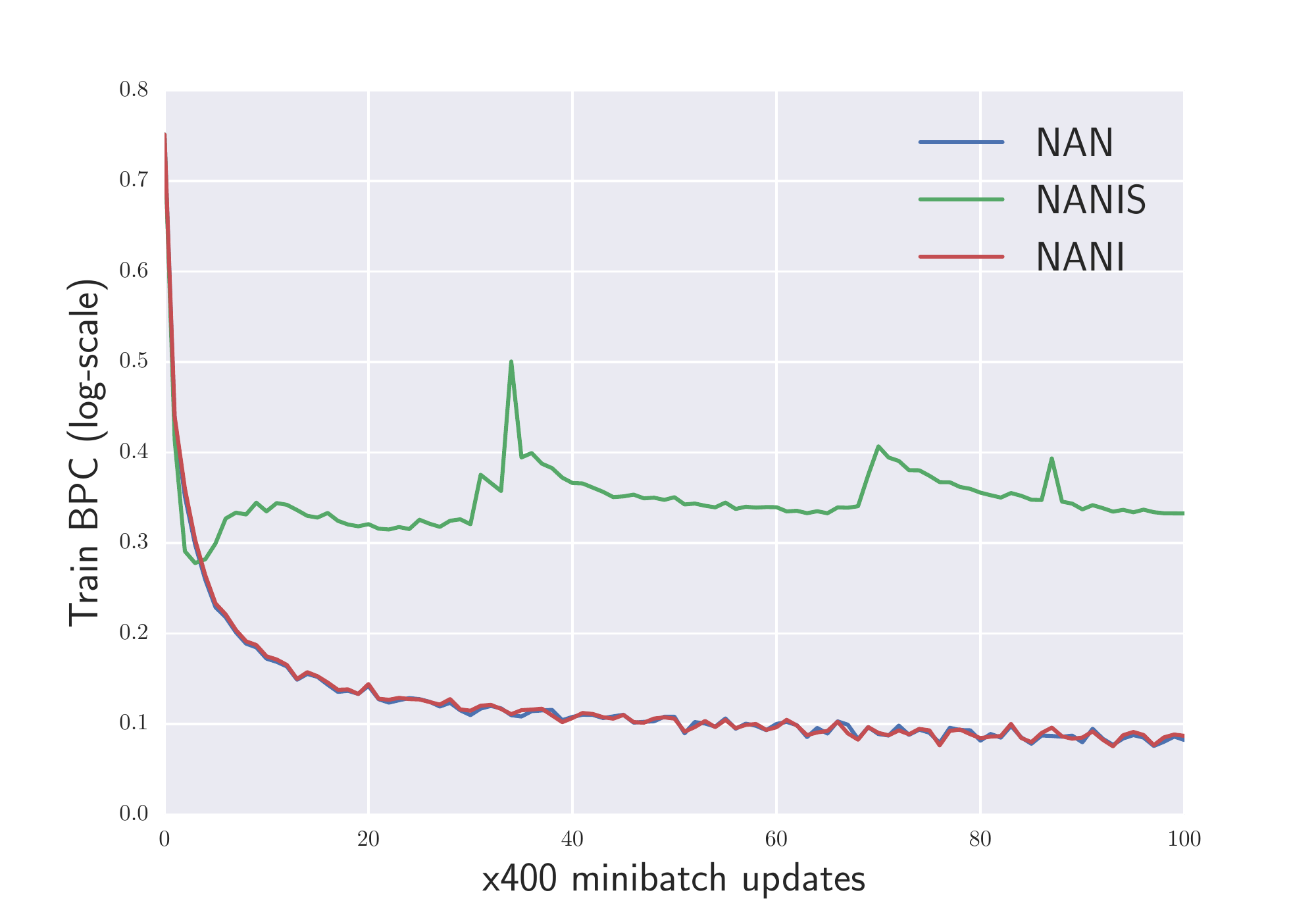}
    \caption{We show the learning curves of a simple character-level 
            GRU language model over the sequences of length $200$ on PTB\@. NANI and NAN, 
            have very  similar learning curves. NANIS in the beginning of the training has a better progress
            than NAN and NANIS, but then training curve stops improving.}
    \label{fig:ptb_learn_curve_chars}
\end{figure}

We further investigated the performances of network with activation functions using NAN, NANI and 
NANIS on penntreebank~(PTB) character-level language modeling. We used a GRU language model over
sequences of length $200$. We used the same model and train all the activation functions with the same
hyperparameters except we ran a grid-search for $\sigma$ for NANI and NANIS from $[1,~0.01]$ with
$8$ values. We choose the best $\sigma$ based on the validation bit-per-character~(BPC). We have 
not observed important difference among NAN and NANI in terms of training performance as seen on
Figure~\ref{fig:ptb_learn_curve_chars}.

\subsection{Learning to Execute}

The problem of predicting the output of a short program introduced in
\cite{zaremba2014learning}~\footnote{The code is residing at \url{https://github.com/wojciechz/learning_to_execute}.
We thank authors for making it publicly available.} proved challenging for modern deep learning
architectures. The authors had to use curriculum learning~\citep{Bengio+al-2009-small} to let 
the model capture knowledge about the easier examples first and increase the level of difficulty of
the examples further down the training.

We replaced all $\sigm$ and $\tanh$ non-linearities in the reference model with their noisy
counterparts. We changed the default gradient clipping to $5$ from $10$ in order to avoid
numerical stability problems. When evaluating a network, the length (number of lines) of the
executed programs was set to $6$ and nesting was set to $3$,
which are default settings in the released code for these tasks. Both the reference model
and the model with noisy activations were trained with ``combined'' curriculum which
is the most sophisticated and the best performing one.

Our results show that applying the proposed activation function leads to better 
performance than that of the reference model. Moreover it shows that the method
is easy to combine with a non-trivial learning curriculum. The results are presented
in Table~\ref{tbl:learning-to-execute} and in Figure~\ref{fig:learning-to-execute}

\begin{table}[t]
\centering
\caption{Performance of the noisy network on the {\em Learning to Execute} /\ task. Just
  changing the activation function to the proposed noisy one yielded about 2.5\% improvement
in accuracy.}
\label{tbl:learning-to-execute}
\begin{tabular}{@{}lr@{}} \midrule
Model name                    					& Test  Accuracy \\ \midrule
Reference Model                        		& 46.45\%          \\
Noisy Network(\textbf{NAH})             	& \textbf{48.09\%}
\end{tabular}
\end{table}

\begin{figure}[ht]
    \centering
    \includegraphics[width=0.98\columnwidth]{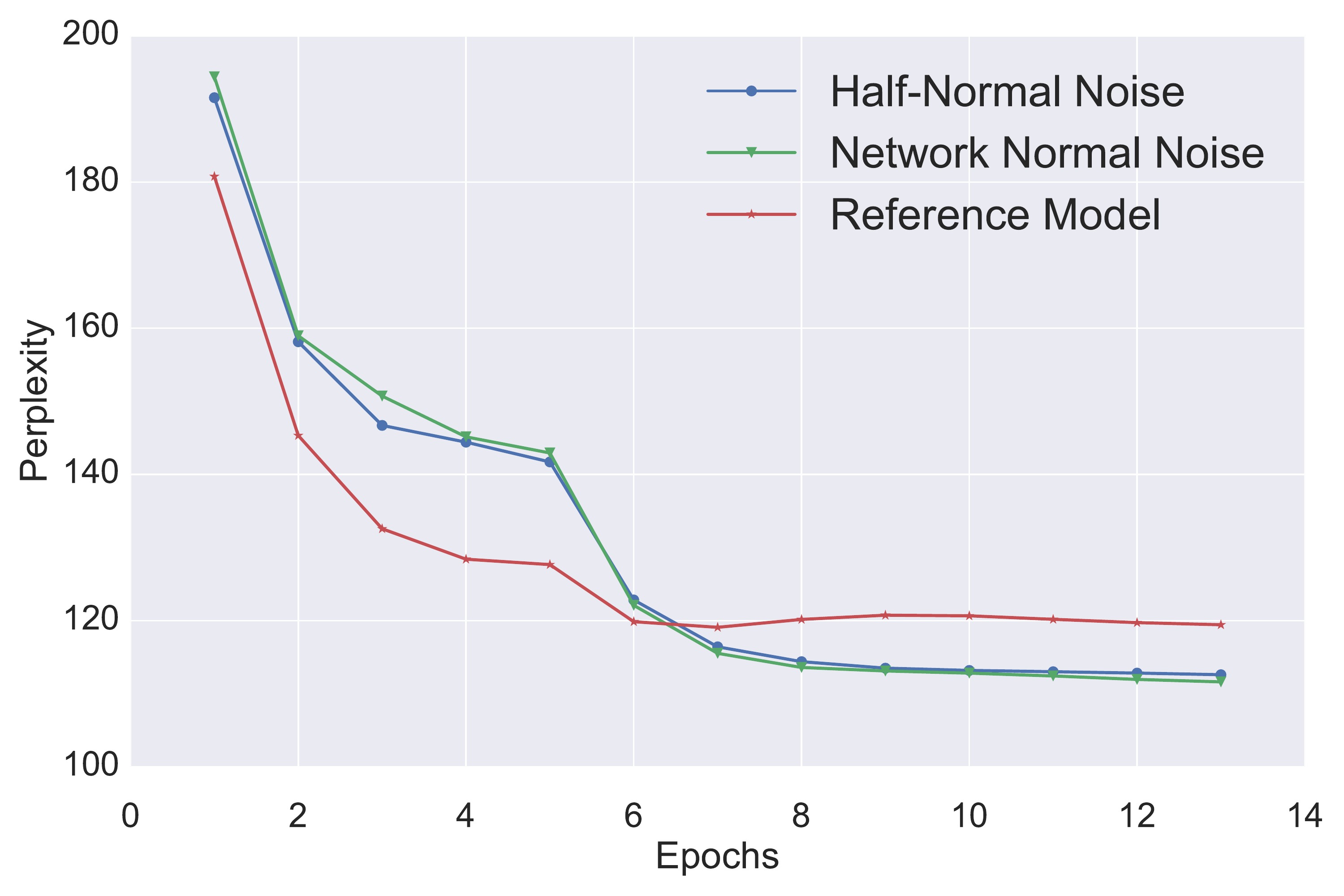}
    \caption{Learning curves of validation perplexity for the LSTM language model on word level on
    Penntreebank dataset.}
    \label{fig:learning-to-execute}
\end{figure}

\begin{figure}[ht]
    \centering
    \includegraphics[width=0.98\columnwidth]{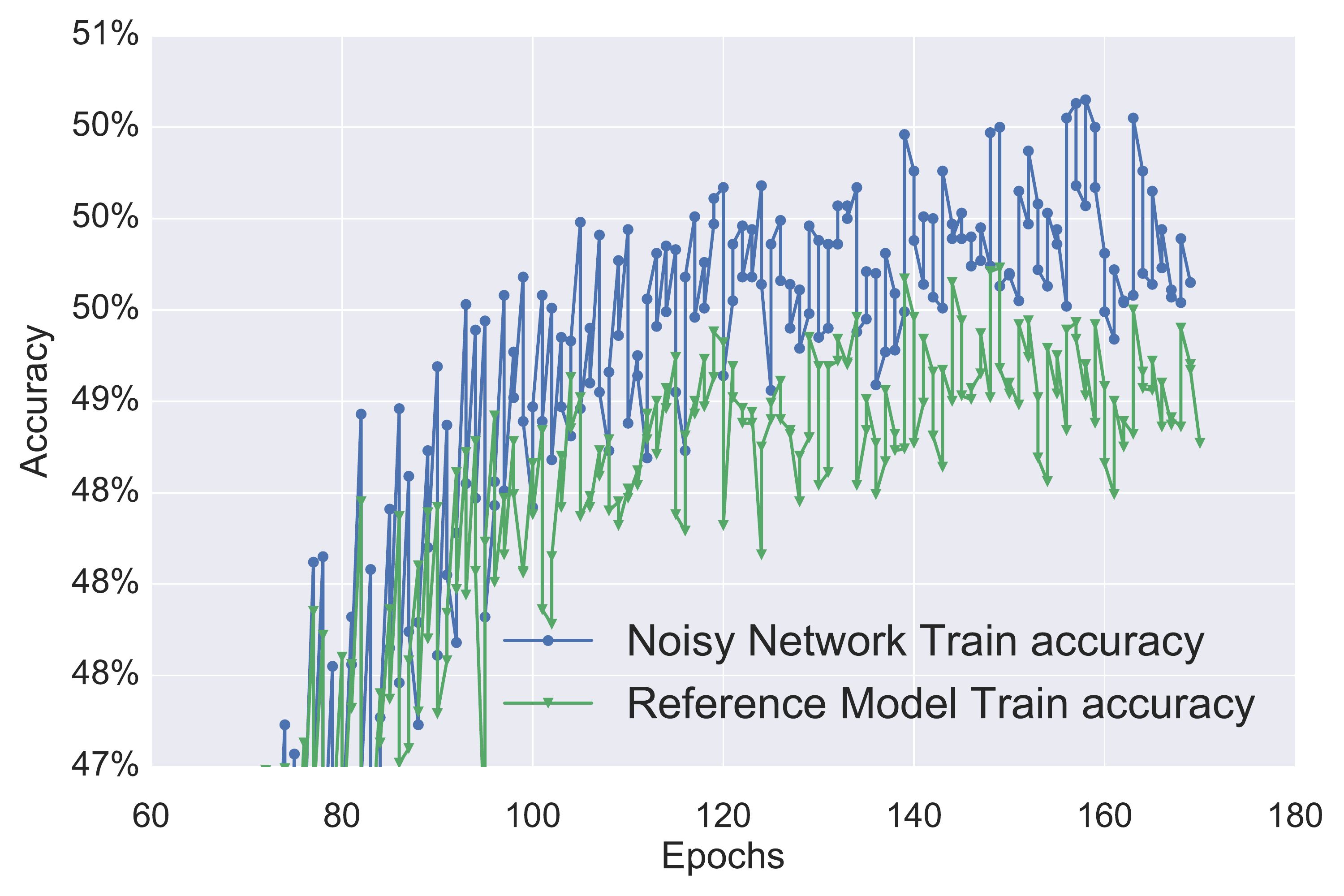}
    \caption{Training curves of the reference model~\citep{zaremba2014learning}
      and its noisy variant on the ``Learning To Execute'' problem. The noisy
      network converges faster and reaches a higher accuracy, showing that the noisy activations
       help to better optimize for such hard to optimize tasks.}
    \label{fig:learning-to-execute}
\end{figure}

\subsection{Penntreebank Experiments}

We trained a $2-$layer word-level LSTM language model on Penntreebank. We used the same
model proposed by \citet{zaremba2014recurrent}.~\footnote{We used the code provided in
\url{https://github.com/wojzaremba/lstm}}We simply replaced all $\sigm$ and $\tanh$ units
with noisy $\hardsigm$ and $\hardtanh$ units. The reference model is a well-finetuned
strong baseline from \cite{zaremba2014recurrent}. For the noisy experiments we used exactly
the same setting, but decreased the gradient clipping threshold to $5$ from $10$.
We provide the results of different models in Table~\ref{tbl:penntreebank_word_ppl}.
In terms of validation and test performance we did not observe big difference between
the additive noise from Normal and half-Normal distributions, but there is a substantial
improvement due to noise, which makes this result the new state-of-the-art on this
task, as far as we know.

\begin{table}[t]
    \centering
    \caption{Penntreebank word-level comparative perplexities. We only replaced in the
      code from~\citet{zaremba2014recurrent} the
      $\sigm$ and $\tanh$ by corresponding noisy variants and observe a substantial
    improvement in perplexity, which makes this the state-of-the-art on this task.}
    \label{tbl:penntreebank_word_ppl}
    \begin{tabular}{@{}lll@{}}
        \toprule
                                           & Valid  ppl & Test  ppl      \\ \midrule
        Noisy LSTM + \textbf{NAN}      & 111.7      & \textbf{108.0} \\
        Noisy LSTM + \textbf{NAH} & 112.6      & \textbf{108.7} \\
        LSTM (Reference)                   & 119.4      & 115.6          \\ \bottomrule
    \end{tabular}
\end{table}

\begin{table*}[htb]
\centering
\caption{Image Caption Generation on Flickr8k. This time we added noisy activations
  in the code from \cite{xu2015show} and obtain substantial improvements on the  higher-order
  BLEU scores and the METEOR metric, as well as in NLL\@. Soft attention and hard attention
  here refers to using backprop versus REINFORCE when training the attention mechanism. We 
  fixed $\sigma=0.05$ for NANI and $c=0.5$ for both NAN and NANIL.
}
\label{tbl:Flickr8kCapgen}
\begin{tabular}{@{}lllllll@{}}\midrule
                                          & BLEU -1 & BLEU-2 & BLEU-3 & BLEU-4 & METEOR & Test NLL \\ \midrule
    Soft  Attention (Sigmoid and Tanh) (Reference)     & \textbf{67}      & 44.8   & 29.9   & 19.5   & 18.9   & 40.33     \\
    Soft Attention ({\bf NAH Sigmoid \& Tanh}) & 66      & \textbf{45.8}   & \textbf{30.69}   & \textbf{20.9}   & \textbf{20.5}   & 40.17 \\ 
    Soft Attention ({\bf NAH Sigmoid \& Tanh} wo dropout) & 64.9      & 44.2 & \textbf{30.7}
    & \textbf{20.9}  & 20.3   & \textbf{39.8} \\ 
    Soft Attention ({\bf NANI Sigmoid \& Tanh}) & 66      & 45.0 & 30.6   &  20.7   & \textbf{20.5}   & 40.0 \\ 
    Soft Attention ({\bf NANIL Sigmoid \& Tanh}) & 66      & 44.6 & 30.1   & 20.0   & \textbf{20.5}   & 39.9 \\ 
    \midrule
    Hard Attention (Sigmoid and Tanh)          & 67      & 45.7   & 31.4   & 21.3   & 19.5   &  -
\end{tabular}
\end{table*}

\subsection{Neural Machine Translation Experiments}

We have trained a neural machine translation (NMT) model on the Europarl dataset with the neural attention model
\cite{bahdanau2014neural}.\footnote{Again, we have used existing code, provided in
\url{https://github.com/kyunghyuncho/dl4mt-material}, and only changed the nonlinearities} We have
replaced all $\sigm$ and $\tanh$ units with their noisy counterparts.
We have scaled down the weight matrices initialized to be orthogonal scaled by multiplying with $0.01$.
Evaluation is done on the {\tt newstest2011} test set. All models are trained with early-stopping.
We also compare with a model with $\hardtanh$ and $\hardsigm$ units and our model using
noisy activations was able to outperform both, as shown in Table~\ref{tbl:NMTEuroparl}.
Again, we see a substantial improvement (more than $2$ BLEU points) with respect to the reference
for English to French machine translation.

\begin{table}[t]
\centering
\caption{Neural machine Translation on Europarl. Using existing code from
  \cite{bahdanau2014neural} with nonlinearities replaced by their noisy versions,
  we find much improved performance (2 BLEU points is considered a lot for machine translation).
We also see that simply using the hard versions of the nonlinearities buys about half of the gain.}
\label{tbl:NMTEuroparl}
\begin{tabular}{@{}lll@{}}
\toprule
                               & Valid  nll & BLEU  \\ \midrule
Sigmoid and Tanh NMT (Reference)    & 65.26      & 20.18 \\
Hard-Tanh and Hard-Sigmoid NMT & 64.27      & 21.59 \\
Noisy (\textbf{NAH}) Tanh and Sigmoid NMT & \textbf{63.46}      & \textbf{22.57}
\end{tabular}
\end{table}

\subsection{Image Caption Generation Experiments}

We evaluated our noisy activation functions on a network trained on the Flickr8k dataset.
We used the soft neural attention model proposed in \cite{xu2015show} as our reference model.\footnote{We used the code provided at~\url{https://github.com/kelvinxu/arctic-captions}.}
We scaled down the weight matrices initialized
to be orthogonal scaled by multiplying with $0.01$.
As shown in Table~\ref{tbl:Flickr8kCapgen}, we were able to obtain better results
than the reference model and our model also outperformed the best model provided in
\cite{xu2015show} in terms of Meteor score.

\citep{xu2015show}'s model was using dropout with the ratio of $0.5$ on the output of the LSTM
layers and the context. We have tried both with and without dropout, as in 
Table~\ref{tbl:Flickr8kCapgen}, we observed improvements
with the addition of dropout to the noisy activation function. But the main improvement seems to
be coming with the introduction of the noisy activation functions since the model without 
dropout already outperforms the reference model.

\vspace*{-1mm}
\subsection{Experiments with Continuation}
\vspace*{-1mm}

We performed experiments to validate the effect of annealing the noise to obtain
a continuation method for neural networks.

We designed a new task where, given a random sequence of integers, the objective is to predict
the number of unique elements in the sequence. We use an LSTM network over the input sequence,
and performed a time average pooling over the hidden states of LSTM to obtain a fixed-size
vector. We feed the pooled LSTM representation into a simple (one hidden-layer) ReLU 
MLP in order to predict the unique number of elements in the input sequence. In 
the experiments we fixed the length of input sequence to $26$ and the input values 
are between $0$ and $10$. In order to anneal the noise, we started training with 
the scale hyperparameter of the standard deviation of noise with $c=30$ and 
annealed it down to $0.5$ with the schedule of $\frac{c}{\sqrt{t+1}}$ where $t$
is being incremented at every $200$ minibatch updates. When noise annealing is combined with a curriculum
strategy (starting with short sequences first and gradually increase the length of the training sequences),
the best models are obtained. 

\begin{table}[t]
\vspace*{-2mm}
\centering
\caption{Experimental results on the task of finding the unique number of elements in a random
  integer sequence. This illustrates the effect of annealing the noise level, turning the training
  procedure into a continuation method. Noise annealing yields better results than the curriculum.}
\label{tbl:unique_number_cont}
\begin{tabular}{@{}ll@{}} \midrule
                                    & Test  Error \% \\ \midrule
    LSTM+MLP(Reference)         & 33.28          \\
    Noisy LSTM+MLP(\textbf{NAN})               & 31.12          \\
Curriculum LSTM+MLP                    & 14.83          \\ \midrule
    Noisy LSTM+MLP(\textbf{NAN}) Annealed Noise & \textbf{9.53} \\
    Noisy LSTM+MLP(\textbf{NANIL}) Annealed Noise & 20.94
\end{tabular}
\vspace*{-2mm}
\end{table}

\begin{figure}[ht]
  \vspace*{-2mm}
    \centering
    \includegraphics[width=0.96\columnwidth]{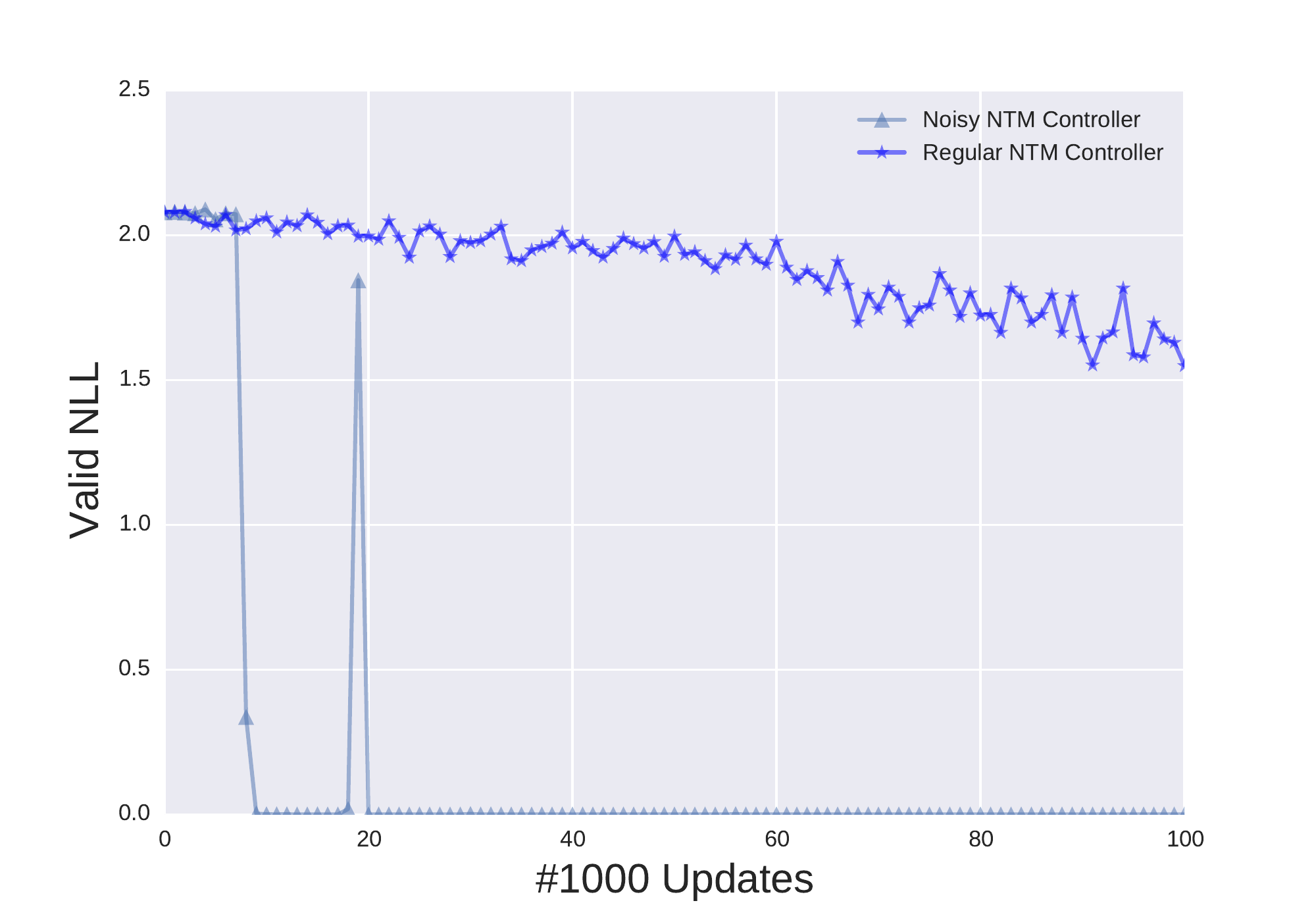}
    \caption{Validation learning curve of NTM on Associative recall task evaluated over items of
    length $2$ and $16$. The NTM with noisy controller converges much faster and solves the task.}
    \label{fig:ntm_annealing}
  \vspace*{-2mm}
\end{figure}

As a second test, we used the same annealing procedure in order to train
a Neural Turing Machine (NTM) on the associative recall task
\cite{graves2014neural}. We trained our model with a minimum of $2$ items and a maximum
of $16$ items. We show results of the NTM with noisy activations in the controller, with annealed noise, and
compare with a regular NTM in terms of validation error. As can be seen in Figure~\ref{fig:ntm_annealing},
the network using noisy activation converges much faster and nails the task, whereas the original network
failed to approach a low error.

\section{Conclusion}

Nonlinearities in neural networks are both a blessing and a curse. A
blessing because they allow to represent more complicated functions and a
curse because that makes the optimization more difficult. For example, we
have found in our experiments that using a hard version (hence
more nonlinear) of the sigmoid and $\tanh$ nonlinearities often improved
results. In the past, various strategies have been proposed to help
deal with the difficult optimization problem involved in training some deep
networks, including curriculum learning, which is an approximate form
of continuation method. Earlier work also included softened versions of
the nonlinearities that are gradually made harder during training.
Motivated by this prior work, we introduce and formalize the concept of 
noisy activations as a general framework for injecting noise in nonlinear 
functions so that large noise allows SGD to be more exploratory. We 
propose to inject the noise to the activation functions either at the input 
of the function or at the output where unit would otherwise saturate, and allow gradients 
to flow even in that case. We show that our noisy activation functions are 
easier to optimize. It also, achieves better test errors, since the noise injected to the 
activations also regularizes the model as well. Even with a fixed noise level, we found
the proposed noisy activations to outperform their sigmoid or $\tanh$ counterpart
on different tasks and datasets, yielding state-of-the-art or competitive results with 
a simple  modification, for example on PennTreebank. In addition, we found that 
annealing the noise to obtain a continuation method could further improved performance.

\bibliographystyle{icml2016}
\bibliography{refs,strings,ml}
\section*{Acknowledgements} 
The authors would like to acknowledge the support of the following agencies for research funding
and computing support: NSERC, Calcul Qu\'{e}bec, Compute Canada, Samsung, 
the Canada Research Chairs and CIFAR.\@ We would also like to thank the developers of
Theano~\footnote{\url{http://deeplearning.net/software/theano/}}, for developing 
such a powerful tool for scientific computing. Caglar Gulcehre also thanks to 
IBM Watson Research and Statistical Knowledge Discovery Group at IBM Research for supporting this 
work during his internship.

\end{document}